\documentclass[sigconf,nonacm]{acmart}
\usepackage{amsmath}
\usepackage{pgfplots}
\usepackage{mathtools}
\usepackage{algorithm}
\usepackage{float}
\usepackage{adjustbox} 
\usepackage{algpseudocode}
\usepackage{graphicx} 
\usepackage{subcaption} 
\usepackage{enumitem} 
\usepackage{booktabs}
\usepackage{multirow} 
\usepackage{tabularx}  
\usepackage{placeins}  
\usepackage{pgfplots}
\usepackage{colortbl}
\usepackage{pgfplotstable}
\pgfplotsset{compat=1.17} 
\usepackage{mfirstuc}
\newcommand{\tmin}[1]{\min_{d\in D_t}\tau(d,#1)}
 
\pgfplotstableread[col sep=comma]{Avg_search_times.csv}\datatable
\pgfplotsset{every bar/.append style={draw=none}}
\pgfplotsset{
  every legend/.append style = { draw=none, fill=none },
  every legend image/.append style = { draw=none }
}

\definecolor{tealDark}{HTML}{00939C}    
\definecolor{tealMedium}{HTML}{6BB5B9} 
\definecolor{tealLight}{HTML}{AAD7D9}    
\definecolor{peachLight}{HTML}{F4B198}   
\definecolor{peachDark}{HTML}{DC7E5A}    
\definecolor{redDark}{HTML}{C22E00}   

\pgfplotsset{%
  /pgf/number format/.cd, 1000 sep={},%
  legend image code/.code={%
    \draw[#1,fill=#1] (0pt,-2pt) rectangle (6pt,2pt);%
  },%
}

\usepgfplotslibrary{groupplots}
\definecolor{baseUncAgn}{RGB}{31,119,180}
\colorlet{colorUncAgn}{baseUncAgn!90!white}

\definecolor{baseCordAgn}{RGB}{255,127,14}
\colorlet{colorCordAgn}{baseCordAgn!90!white}

\definecolor{baseCordAprox}{RGB}{44,160,44}
\colorlet{colorCordAprox}{baseCordAprox!90!white}

\definecolor{baseCordOracle}{RGB}{214,39,40}
\colorlet{colorCordOracle}{baseCordOracle!90!white}

\AtBeginDocument{%
  }

\begin{document}

\title{Reducing Street Parking Search Time via Smart Assignment Strategies}
\subtitle{\textcolor{blue}{\small \textit{*Please cite the SIGSPATIAL’25 version of this paper.}}}

\author{Behafarid Hemmatpour}
\affiliation{%
  \institution{IMDEA Networks Institute
  Universidad Carlos III de Madrid}
  \city{Madrid}
  \country{Spain}}
\email{behafarid.hemmatpour@imdea.org}

\author{Javad Dogani}
\affiliation{%
 \institution{IMDEA Networks Institute}
  \city{Madrid}
  \country{Spain}}
\email{javad.dogani@imdea.org}

\author{Nikolaos Laoutaris}
\affiliation{%
  \institution{IMDEA Networks Institute}
  \city{Madrid}
  \country{Spain}}
\email{nikolaos.laoutaris@imdea.org}

\newcommand{\participants}{participants}
\newcommand{\participant}{participant}
\newcommand{\competitors}{competitors}
\newcommand{\competitor}{competitor}

\newcommand{\Participants}{Participants}
\newcommand{\Participant}{Participant}
\newcommand{\Competitors}{Competitors}
\newcommand{\Competitor}{Competitor}

\newcommand{\partc}{partc} 
\newcommand{\comp}{comp}  

\newcommand{\uncAgn}{\textit{Unc-Agn}}
\newcommand{\cordAgn}{\textit{Cord-Agn}}
\newcommand{\cordOracle}{\textit{Cord-Oracle}}
\newcommand{\cordApprox}{\textit{Cord-Approx}}

\begin{abstract}
In dense metropolitan areas, searching for street parking adds to traffic congestion. Like many other problems, real-time assistants based on mobile phones have been proposed, but their effectiveness is understudied. This work quantifies how varying levels of user coordination and information availability through such apps impact search time and the probability of finding street parking. Through a data-driven simulation of Madrid's street parking ecosystem, we analyze four distinct strategies: uncoordinated search (\uncAgn), coordinated parking without awareness of non-users (\cordAgn), an idealized oracle system that knows the positions of all non-users (\cordOracle), and our novel/practical {\cordApprox} strategy that estimates non-users' behavior probabilistically. The {\cordApprox} strategy, instead of requiring knowledge of how close non-users are to a certain spot in order to decide whether to navigate toward it, uses past occupancy distributions to elongate physical distances between system users and alternative parking spots, and then solves a Hungarian matching problem to dispatch accordingly. In high-fidelity simulations of Madrid’s parking network with real traffic data, users of {\cordApprox} averaged 6.69 minutes to find parking, compared to 19.98 minutes for non-users without an app. A zone-level snapshot shows that {\cordApprox} reduces search time for system users by 72\% (range = 67–76\%) in central hubs, and up to 73\% in residential areas, relative to non-users.
\end{abstract}


\maketitle
\section{Introduction}
Despite significant advances in vehicular navigation systems, the persistent challenge of locating available street parking spots remains largely unaddressed. In general, drivers spend a substantial amount of time annually, often dozens of hours, looking for parking at their destinations~\cite{shoup2007cruising}. This exacerbates multiple systemic issues of densely populated areas, such as heightened traffic, increased vehicular emissions, unnecessary time expenditure, elevated driver frustration, stress, and risk of road accidents. Previous research has looked at the street‑parking problem only sporadically. Some experiment with the amount of information available to drivers via an intelligent parking system, showing that real-time availability data can reduce search time~\cite{dalla2022providing, rodriguez2024analysis}.  Other studies examine how drivers compete for parking spots, often relying on game-theoretic or experimental models to simulate strategic decision-making and the effects of pricing or reservations~\cite{Tan2024, sun2024pricing}. Another line of work focuses on analyzing parking conditions in particular places, using city-specific data to evaluate how demand, land use, and local policies shape curbside occupancy~\cite{millard2014curb, qin2022analysis, Wu2022}. However, no prior work brings all of the above elements together. Hence, in this study, we split drivers into two groups, one that uses a smart street-parking application providing guidance and routes to available street parking spots (henceforth referred to as \participants{}) and another composed of uninformed non-user drivers who search for street parking without system assistance (henceforth referred to as \competitors{}). With finite parking spots, benefits to \Participants{} necessarily diminish availability for \competitors{}. We vary the levels of coordination and information available to \participants{} and evaluate their impacts on street parking search time and success ratio (Section~\ref{results}) under realistic parking conditions, assuming a fixed adoption rate.

We propose and evaluate four distinct smart parking allocation strategies under varying levels of information and coordination among our \participants{}. 1) Our internal baseline among app users, \textit{Uncoordinated Agnostic} (\uncAgn{}), in which \participants{} obtain system-wide parking availability but navigate independently toward their nearest available spots, while \competitors{} search without system assistance. 2) The \textit{Coordinated Agnostic} (\cordAgn) strategy, in which our \participants{} coordinate to avoid contention over parking spots, but are agnostic to competing agents that are outside our system (\competitors{}). 3) The \textit{Coordinated Oracle} (\cordOracle) strategy, in which the system has perfect knowledge of all parking spots and the current positions of \participants{} and \competitors{}. 
This allows the system to avoid assignments likely to be preempted by closer \competitors{}, minimizing wasted trips. This unrealistic, perfect-awareness setting is included solely as an upper-bound benchmark. 4) \textit{The Coordinated Approximate} (\cordApprox{}) strategy, which is a practical realization of the above oracle strategy leveraging historical data to probabilistically estimate the impact of the presence of \competitors{} without requiring knowledge of their exact locations. For comparability, the numbers of \participants{} and \competitors{} are fixed across all strategies. Our evaluation focuses on search time and success ratio, leaving network-wide congestion to future work. For cross-group comparisons to current practice, \competitors{} serve as the real-world baseline, while Strategy \uncAgn{} is the app-user baseline (information only, no coordination).

\textbf{Our Contributions}: This work makes four key contributions to intelligent urban parking systems: First, we quantify the performance benefits of coordinated street parking allocation under varying information conditions. Second, we analyze design trade-offs, including the value of coordination among \participants{} and the impact of awareness of \competitors{} on system efficiency. Third, we develop a comprehensive evaluation framework using realistic parameters: actual traffic-intensity data, a Geohash-aligned grid of Madrid, zone-specific parking capacities, historical occupancy patterns, and empirically calibrated parking duration/turnover. Fourth, we propose a practical forecasting approach that approximates an oracle's knowledge of \competitors{} current positions. Our methodology combines real-world traffic intensity data with behavioral modeling to bridge the gap between theoretical parking optimization and practical implementation constraints.
 
\textbf{Our findings:} Using real-world data from over 2 million observations across Madrid, Spain~\cite{Dataset}, our experimental evaluation offers key insights into how intelligent parking systems perform and influence driver behavior in realistic urban settings:
\\
\noindent $\bullet$ Smart-parking systems provide diminishing benefits at the extremes, when parking spot availability is high (40–45\%) or very low (0–5\%). In the intermediate range (20–25\% availability), however, smart-parking systems have strong potential to improve street parking performance. We show that having hints about free street spots (i.e., \uncAgn{}) is insufficient: contention over these spots can erode the advantage of \participants{} over \competitors{} and, in extreme cases, even result in worse performance than \competitors{}.
\\
\noindent $\bullet$  Introducing coordination among our \participants{}, through \cordAgn{}, \cordApprox{}, and \cordOracle{}, leads to substantial improvements in both parking success ratio and reduction in search time.
\\
\noindent \textbf{Parking Success Ratio.} Starting with the \uncAgn{} strategy, our \participants{} achieve a 34.25\% success ratio, trailing the \competitors{}' 38.63\% by about four percentage points. Introducing coordination (\cordAgn{}) lifts our \participants{}' performance sharply to 70.04\%, while \competitors{} remain at 30.71\%. With the full-information \cordOracle{}, where the positions of all \competitors{} are known, \participants{} attain 85.32\%, far above the \competitors{}' 27.76\%. Finally, under the practical oracle approximation (\cordApprox{}), \participants{} reach 77.54\%, compared with 28.71\% for the \competitors{}. Overall, coordination and predictive awareness increase our \participants{}' parking success ratio by more than double relative to \uncAgn{}, an advantage of up to 50 percentage points (pp) over \competitors{}.
\\
\noindent \textbf{Average Search Time.} \cordApprox{} also excels in time efficiency \emph{for successful parking attempts}, reducing average search time to 6.69 minutes, compared to 19.98 minutes for \competitors{}. \cordOracle{} averages 8.83 minutes, while \cordAgn{} averages 9.86 minutes. Notice that it should not come as a surprise that \cordApprox{} has a smaller search time than \cordOracle{}. This is because we compute this metric only for successful parking attempts, and the two strategies have a different success ratio, so the search times are not directly comparable. Overall, looking at both metrics, it is clear that \cordApprox{} offers superior performance compared to \competitors{}, while remaining close to the impractical \cordOracle{} upper bound.
\\
\noindent $\bullet$ Zooming in on individual neighborhoods in central Madrid reveals that the benefits of our \cordApprox{} strategy are even more pronounced than the encouraging reduction in the city-wide mean of 66.5\%. \cordApprox{} reduces search time for \participants{} by 76\% in \textbf{Culture \& Transport Hubs}, 72\% in the mixed \textbf{Residential \& Light Industry} belt, and 73\% in the \textbf{Traditional Residential} zone. The strategy not only improves city-wide performance but excels even further at the neighborhood level, turning a global parking challenge into a series of localized advantages that ease the search time across diverse traffic conditions.

\section{Related Work}
We can break down street‑parking research efforts into three main streams: (i) \emph{information‑centric guidance systems}, (ii) \emph{coordinated allocation strategies}, and (iii) \emph{empirical field studies}. Although each strand is well developed, the literature lacks a metropolitan‑scale, data‑driven comparison across them. Our study addresses this gap.

\textbf{Information‑centric guidance systems.} Sensor‑based systems such as \emph{ParkNet}~\cite{mathur2010parknet}, recent sensing approaches~\cite{pitt2023driveby}, and comparative evaluations of fixed and mobile sensing~\cite{roman2018detecting} demonstrate how drive‑by or crowdsourced sensing can publish real‑time occupancy of street maps with high accuracy. More recent work uses parking-meter transaction streams and sparse curbside sensors as a city-wide probe network to forecast block-level vacancy rates without full instrumentation~\cite{kotb2016iparker}. Integrating richer context improves forecasts: Inam et al.~\cite{inam2022multisource} integrate occupancy, weather, pedestrian count, and traffic data streams to train a Random Forest model that predicts on-street parking vacancy 5–10 minutes in advance with 81\% accuracy, achieving a 10 percentage point improvement over models using an occupancy baseline. Hybrid platforms such as D2Park~\cite{Zhao2021D2Park} combine real-time sensing with user preferences (e.g., walking distance, cost) to dynamically guide drivers toward likely available spots, reducing average search time by over 20\%. Although these methods greatly improve individual awareness, they typically do not resolve contention among multiple informed drivers.

\textbf{Coordinated allocation strategies.} A complementary line of work assigns drivers to spots. A Hungarian‑based optimal matching~\cite{zhao2014algorithm} reduces search time by roughly $50\%$ in simulations. A recent study~\cite{sayarshad2020scalable} propose a scalable, non-myopic atomic game that jointly optimises assignment and dynamic pricing, raising social welfare by up to 54\% on San Francisco data. Tan et al.~\cite{Tan2024} proved, in a congestion‑game model, that advance reservation strictly dominates blind search. However, most coordination studies benchmark against oracle upper bounds in synthetic settings and assume perfect compliance, conditions rarely achievable in practice.

\textbf{Empirical field studies.} Large-scale deployments are rare. The SFpark pilot~\cite{zimmerman2014san}, observed a 43\% drop in average searching time after introducing demand-based pricing and real-time guidance. Barcelona’s on-street sensor rollout showed how sparse instrumentation plus predictive analytics can reduce cruising under real conditions~\cite{zambanini2020detection}. A study in a European urban center used a survey-based approach to analyze how pricing reform influenced occupancy and driver behavior~\cite{cats2016survey}. These studies highlight the complexities of real-world deployments, where sensor coverage, driver compliance, and fluctuating demand collectively influence outcomes.

\textbf{Positioning of this study.} Using street occupancy records for Madrid~\cite{Dataset}, we present a unique city‑scale empirical comparison of four strategies: \emph{Uncoordinated Agnostic} (\uncAgn{}), \emph{Coordinated Agnostic} (\cordAgn{}), \emph{Coordinated Approximate} (\cordApprox{}), and \emph{Coordinated Oracle} (\cordOracle{}). Our experiments (i) quantify the marginal benefits of uncoordinated and coordinated strategies, (ii) show that modest coordination captures most oracle gains, and (iii) introduce a practical approximation (\cordApprox{}) that approaches oracle performance without tracking \competitors{} in real time. By empirically bridging the gap between theoretical optima and deployable systems, our findings offer concrete guidance for designing future smart street‑parking deployments.

\section{Problem Formulation}
This study examines two core questions about smart parking systems: (1) How do key design choices (coordination, information granularity) affect real traffic? (2) How do environmental factors (observability radius, spot availability, traffic intensity) impact performance? Both require analyzing the interaction between information and coordination. To rigorously analyze the impact of smart parking strategies in real deployments, we built a city-scale simulation of Madrid's curbside ecosystem. We (1) simulate parking supply, demand, and traffic dynamics under controlled conditions, and (2) isolate key factors, information granularity, coordination, and adoption rate, to assess their effects on system performance. 

\subsection{Definitions}
We provide below the formal definitions of the core symbols used in our simulation-based study (summarized in Table~\ref{tab:notation}). Further details on the simulation findings are presented in Section~\ref{design}.
\newcolumntype{L}{>{\raggedright\arraybackslash}X}
\begin{table}
  \small    
  \setlength\tabcolsep{4pt}
  \caption{Table of Notation}
  \label{tab:notation}
  \begin{tabularx}{\columnwidth}{@{}lL@{}}
    \toprule
    \textbf{Symbol} & \textbf{Description} \\
    \midrule
    $n$ & grid dimension; city is an $n\times n$ mesh. \\
    $z = (i,j)$ & generic grid cell with indices $i,j \in \{0,1,\dots,n-1\}$. \\
    $T$ ($t\!\in\!T$) & discrete simulation horizon; $t$ is a step (one step = one tick). \\
    $\mathcal{D}$, $D_t$ & all \participants{} (app-enabled drivers); $D_t$ is the subset active at time $t$. \\
    $\mathcal{C}$, $C_t$ & all \competitors{} (non-users); $C_t$ is the subset active at time $t$. \\
    $t_{max}$ & search-time budget; attempts exceeding $t_{max}$ are failures.\\
    $b_{i,j}$ & spot (bay) capacity in cell $(i,j)$. \\
    $B$ & total spot capacity in the city, i.e., $\sum_{i,j}b_{i,j}$. \\
    $\mathcal{S}_t$ & set of available spots at time $t$.\\
    $R$ & \competitor{} observability radius (in grid cells). \\
    $\tau(d,s),\;\tau(c,s)$ & Manhattan distance (travel time) from \participant{} $d$/\competitor{} $c$ to spot $s$. \\
    $x_{d,s,t}\!\in\!\{0,1\}$ & binary decision variable that equals 1 if \participant{} $d$ is assigned to spot $s$ at time $t$. \\
    $\mathcal{A}^{\comp}_t$ & all spots captured by \competitors{} at time $t$. \\
    $\mathcal{A}^{\partc}_t$ & all spots assigned to \participants{} at time $t$.  \\
    $t_c$ & number of steps available for \competitor{} c before \participant{} $d$ comes within one step of the visibility radius $R$ around spot $s$.\\
    $C^{*}_{d,s,t}$ & \competitors{} closer than \participant{} $d$ to spot $s$ but still outside radius $R$.\\
    $p(c,s)$ & probability \competitor{} $c$ reaches spot $s$ first. \\
    $\mathcal N_c^{\le t_c}$ & cells \competitor{} $c$ can reach within $\le t_c$ steps. \\
    $s^{*}_{d}$ & closest spot chosen by \participant{} $d$. \\
    $\mathcal{H}$ & historical corpus of spot parking success ratios. \\
    $g_k$  & Geohash label representing the grid cell $z$ at position $(i,j)$; $k$ indexes that cell. \\
    $\rho_{k,t}$ & observed parking success ratio in Geohash $g_k$ at time $t$. \\
    $\hat{p}_{k(s),t}$ & predicted spot availability probability (Ridge regression). \\
    $\tilde{\tau}(d,s)$ & effective distance defined as $\tau(d,s)/\hat{p}_{k(s),t}$, at time $t$.\\
    $\mathbf{M}_t$ & cost matrix at time $t$. \\
    $\mathbf{M}_t(d, s)$ &  entry $(d,s)$ of the cost matrix $\mathbf M_t$.\\ 
    $k(s)$ & mapping from spot $s$ to its Geohash index $k$. \\
    $\mathbf{X},\mathbf{y}$ & feature matrix and target vector for Ridge model. \\
    $\beta,\;\lambda$ & coefficients and L2 regularization weight in Ridge regression. \\
    \bottomrule
  \end{tabularx}
\end{table}
\\
\noindent $\bullet$ The city is divided into an $n \times n$ grid, indexed by $(i, j)$ where $i, j \in \{0, 1, \ldots, n-1\}$. The simulator evolves over discrete time steps $t\in T$.
\\
\noindent $\bullet$ Let $D_t \subseteq \mathcal{D}$ be the set of \participants{} (app-enabled drivers) active at time $t$, and $C_t \subseteq \mathcal{C}$ the set of \competitors{} (non-users searching without system assistance). Parking infrastructure is defined by a spatial distribution $b_{i,j}\in\mathbb{Z}^+$ indicating spots (bays) capacity in grid cell $z = (i,j)$, with total capacity $B = \sum_{i=0}^{n-1}\sum_{j=0}^{n-1} b_{i,j}$. The system maintains perfect knowledge of street parking spot availability through deployed sensors (see Section~\ref{design}).
\\    
\noindent $\bullet$ The travel time $\tau(d,s)$ is the Manhattan distance between the \participant{} $d$ and spot $s$ (via their Geohash cells); see Section~\ref{design}.
\\
\noindent $\bullet$ The observability radius $R$ is maximum distance within which \competitors{} detect spots. \Competitor{} $c\in C_t$ observes spot $s\in\mathcal{S}_t$ if $\tau(c,s) \leq R$, where $\tau(c,s)$ is Manhattan distance between $c$ and $s$.
\\
\noindent $\bullet$ A parking attempt by an agent is \emph{successful} if the agent reaches the parking spot before anyone else (\participants{} or \competitors{}) and within the per-attempt budget $t_{max}$. Attempts exceeding $t_{max}$ are counted as failures and the agent exits the simulation; real-world recourse (e.g., off-street or farther parking) is out of scope.
\\
\noindent $\bullet$ We extend the parking spot allocation problem to a dynamic setting, where agents and spots appear and disappear over a predefined simulation horizon $T$. 
     The binary decision variable \(x_{d,s,t}\in\{0,1\}\) equals~1 when \participant{} \(d\in D_t\) is assigned to spot \(s\in\mathcal{S}_t\) at time \(t\). Let \(\mathcal{A}^{\comp}_t\subseteq\mathcal{S}_t\) be the set of
      spots already secured by \competitors{} at time~\(t\):
    \begin{equation}
    \begin{split}
    \mathcal{A}^{\comp}_t = \bigl\{\, s \in \mathcal{S}_t \;\big|\; 
        & \exists\, c \in C_t : \tau(c,s) \le R \;\land \\
        & \quad \tau(c,s) < \tmin{s} \,\bigr\}
    \end{split}
    \label{eq:comp-capture}
    \end{equation}
    
      If multiple agents, \participants{}/\competitors{} are equally close to a spot \( s \) or both at the spot \( s \), we break ties uniformly at random; This ensures fairness and avoids systematic bias. If the selected agent is \competitor{} c, the spot s is added to the set \( \mathcal{A}^{\comp}_t \), as defined in Equation~\ref{eq:comp-capture}. The dispatcher then minimizes the current assignment cost over eligible pairs while respecting \competitor{} preemption:

      \begin{equation}
      \min_{x}\;
      \sum_{t\in T}\sum_{d\in D_t}\sum_{s\in\mathcal{S}_t\setminus\mathcal{A}^{\comp}_t}
      \tau(d,s)\,x_{d,s,t},
      \label{eq:dyn-objective}
      \end{equation}
      subject to
      \begin{align}
      &\sum_{s\in\mathcal{S}_t} x_{d,s,t} \;\le\; 1
      &&\forall d\in D_t,
      \label{eq:driver-excl}\\[4pt]
      &\sum_{d\in D_t} x_{d,s,t} \;\le\;
        \mathbb{I}\!\bigl[s\notin\mathcal{A}^{\comp}_t\bigr]
      &&\forall s\in\mathcal{S}_t,
      \label{eq:spot-availability}
      \end{align}
      where \(\mathbb{I}[\cdot]\) is the indicator function. Constraint~\eqref{eq:driver-excl} enforces \participant{} exclusivity, whereas Constraint~\eqref{eq:spot-availability} blocks allocation of spots already captured by \competitors{}.  We solve this problem online by recomputing a Hungarian bipartite match at every \(t\in T\) with updated \participants{}/\competitors{} current positions and spot availability.  Unassigned \participants{} propagate to \(t{+}1\) while moving toward their previous targets, yielding a rolling-horizon scheme that captures (1) real-time occupancy changes, (2) continuous agents' movement, and (3) fluctuating demand.

\subsection{Strategies}
We evaluate four distinct parking allocation strategies, each representing a different level of situational awareness and coordination among \participants{}. Together, they isolate how information and coordination affect outcomes for app users and non-users. For cross-group comparisons to current practice, \competitors{} (blind search) are the real-world baseline; Strategy~1 is an app-user baseline (information only, no coordination).

\subsubsection{Strategy 1: Uncoordinated Agnostic (\uncAgn{})}
In this strategy, each \participant{} $d \in D_t$ has full visibility of all available spots $\mathcal{S}_t$ at time $t$ (information only; no coordination among \participants{}). Each \participant{} $d \in D_t$ independently selects their nearest unoccupied spot:

\begin{equation}
s^{*}_d = \arg\min_{s \in \mathcal{S}_t} \tau(d, s)
\end{equation}
where $s^{*}$ is \participant{} $d$'s target spot. Due to uncoordinated decision-making, multiple \participants{} may target the same spot, resulting in conflicts. Failed assignments trigger renewed search behavior, introducing inefficiencies from contention and redundant travel (Algorithm~\ref{alg:unc-agn}).

\begin{algorithm}
  \caption{Uncoordinated Agnostic (\uncAgn{})}
  \Description{Greedy assignment of each \participant{} to the nearest available parking spot at time step $t$.}
  \begin{algorithmic}[1]
    \State \textbf{Input:} $D_t$~—~\participants{} active at step $t$ (incl. unparked \participants{} from $t-1$); 
           $\mathcal{S}_t$~—~available spots at step $t$; 
           $\{\tau(d,s)\}_{d\in D_t, s\in\mathcal{S}_t}$~—~all the travel times of \participants{} $d$ to all the spots $s$ 
    \State \textbf{Output:} $\mathcal{A}^{\partc}_t$~—~assigned spots to $D_t$
    \State $\mathcal{A}^{\partc}_t= []$
    \For{$d \in D_t$}
        \State $s^{*}_{d} \gets \underset{s \in \mathcal{S}_t}{\arg\min}\;\tau(d, s)$
        \State $\mathcal{A}^{\partc}_t$.\textbf{Add}($s^*_{d}$)
    \EndFor
    \State \textbf{return} $\mathcal{A}^{\partc}_t$
  \end{algorithmic}
  \label{alg:unc-agn}
\end{algorithm}

\subsubsection{Strategy 2: Coordinated Agnostic (\cordAgn{}) }
Here, the global orchestrator knows all free spots and receives \participants{}’ current positions, but not \competitors{}’ positions. At each time step $t$, \participants{} report their positions to a global orchestrator. Then, the global orchestrator constructs a cost matrix $\mathbf{M}_t \in \mathbb{R}^{|D_t| \times |\mathcal{S}_t|}$. 
Each cost entry $\mathbf{M}_t(d, s) \in \mathbf{M}_t$ represents the travel time $\tau(d, s)$ that a \participant{} $d$ needs to take for reaching the spot $s$. 
The orchestrator then solves an optimization problem using $\mathbf{M}_t$ to assign each spot $s_d^*$ to a single \participant{} $d$ with minimum cost on $d$. 

This optimization problem reduces to an assignment problem that minimizes total travel time. The assignment problem is then solved by deploying the Hungarian algorithm~\cite{kuhn1955hungarian} whose time complexity is $O\bigl(|D_t|^3\bigr)$.  This approach eliminates contention among our own \participants{} but may still fail to secure spots by ignoring the effect of non-participating \competitors{} (See Algorithm~\ref{alg:cord-agn}).

\begin{algorithm}
\caption{Coordinated Agnostic (\cordAgn{})}
\label{alg:cord-agn}

\begin{algorithmic}[1]
   \State \textbf{Input:} $D_t$~—~\participants{} active at step $t$ (incl. unparked \participants{} from $t-1$); 
           $\mathcal{S}_t$~—~available spots at step $t$; 
           $\{\tau(d,s)\}_{d\in D_t, s\in\mathcal{S}_t}$~—~all the travel times of \participants{} $d$ to all the spots $s$ 
    \State \textbf{Output:} $\mathcal{A}^{\partc}_t$~—~assigned spots to $D_t$
    \State $\mathcal{A}^{\partc}_t= []$

\State \textbf{Initialize} $\mathbf{M}_t(D_t,\mathcal{S}_t)$
\For{$d \in D_t$}
    \For{ $s \in \mathcal{S}_t$}
        \State $\mathbf{M}_t(d,s) := \tau(d, s)$
    \EndFor
\EndFor
\State $\mathcal{A}^{\partc}_t = \mathbf{HungarianAssign}$($\mathbf{M}_t$) \Comment{\textcolor{blue}{Solve the minimum cost assignment problem with Hungarian algorithm}}
\State \textbf{return} $\mathcal{A}^{\partc}_t$
\end{algorithmic}
\end{algorithm}

\subsubsection{Strategy 3: Coordinated Oracle (\cordOracle{})}
This strategy, denoted by \cordOracle{}, assumes perfect global knowledge: current positions of all \participants{}, all available spots, and all \competitors{} in real time $t$ as shown by Algorithm~\ref{alg:cord-oracle}. This infeasible setting is used only as an upper-bound benchmark. To compute \(\mathbf{M}_t(d, s)\) (the $\mathbf{CalculateCost}$ procedure in Algorithm~\ref{alg:cord-oracle}), we must consider three conditions as specified by Equation~\ref{oracle}, which are detailed below:
\\
\noindent $\bullet$ \textbf{Condition 1:} A \participant{} $d$ takes the spot $s$ if $d$ is closer to $s$ than all the \competitors{} ($\tau(d,s) < \min\limits_{c\in C_t} \tau(c,s)$). In this case, $\mathbf{M}_t(d, s)$ is $\tau(d,s)$ since it takes $\tau(d,s)$ for $d$ to reach to $s$. 
\\
\noindent $\bullet$ \textbf{Condition 2:} A \competitor{} $c$ takes $s$ if any \competitor{} is within the visibility radius $R$ and closer to $s$ than any \participant{} is ($\exists\, c\in C_t\ | \tau(c,s) < \tau(d,s) \wedge \tau(c,s) \le R$). Now, the chance of $s$ being assigned to $d$ is effectively zero, which means that $\mathbf{M}_t(d, s)$ is too high; thus, we determine $\mathbf{M}_t(d, s)$ as infinity.
\\
\noindent $\bullet$ \textbf{Condition 3:} In this last condition, if $c$ is closer than $d$ but still outside $R$ ($\tau(c,s)>R$), $\mathbf{M}_t(d,s)$ depends on a distance-weighted probability. This probability, denoted by $p(c,s)$, is the probability of \competitor{} $c$ being able to reach a range $R$ of spot $s$ before \participant{} $d$. The cost matrix adapts dynamically using a piecewise distance-based function.

\begin{equation}
\begin{aligned}
{\small
\mathbf{M}_t(d, s)= 
\begin{cases}
\tau(d,s), &  \text{if } \tau(d,s) < \min\limits_{c\in C_t} \tau(c,s) \\[4pt]
\infty, &  \text{if } \exists\, c\in C_t\ \text{s.t. } \\
& \quad  \tau(c,s) < \tau(d,s) \\
& \quad \text{and } \tau(c,s) \le R \\[4pt]
\tau(d,s) + \sum\limits_{c \in C^{*}_{d,s,t}} \tau(d,s) \cdot p(c,s), & \text{otherwise}
\end{cases}}
\end{aligned}
\label{oracle}
\end{equation}
Here, \(C^{*}_{d,s,t} = \{\,c \in C_t: \tau(c,s) < \tau(d,s)\ \wedge\ \tau(c,s) > R\}\) denotes the set of \competitors{} who are closer to spot \(s\) than \participant{} \(d\), yet lie outside the visibility radius \(R\). For any such \competitor{} $c$, let
\begin{equation}
\label{eq:tc}
t_c \;=\; \min\bigl(\tau(d,s)-R-1,\;R\bigr)
\end{equation}
be the number of steps available before \participant{} $d$ comes to within one step of the visibility radius $R$ around spot $s$. Define the reachable set
\begin{equation}
\label{eq:Nc}
\mathcal{N}_c^{\le t_c}
  \;=\;
  \bigl\{\,z\in\mathbb Z^{2} : \tau(c,z)\le t_c\bigr\},
\end{equation}
where $z=(i,j)$ denotes a generic grid cell.  Each element of $\mathcal{N}_c^{\le t_c}$ is a location that $c$ could occupy after at most $t_c$ uniform–random moves (one step north, south, east, or west per tick). We define the distance-weighted probability,
\begin{equation}
\label{eq:pcs}
p(c,s)
  \;=\;
  \frac{
    \bigl|\{\,z\in\mathcal{N}_c^{\le t_c} :
            \tau(z,s)=R\}\bigr|
  }{
    \bigl|\mathcal{N}_c^{\le t_c}\bigr|
  }.
\end{equation}
Since Manhattan distance on our grid corresponds directly to travel time, \(t_c\) denotes the \competitor{}’s total time budget before \participant{} $d$ comes to within one step of the visibility radius $R$ around spot $s$. Assuming each move direction at every tick is chosen uniformly at random, all endpoints in \(\mathcal{N}_c^{\le t_c}\) are equally likely. A favorable outcome occurs precisely when that endpoint lands on the radius $R$ circle around spot~$s$, i.e., $\tau(z,s)=R$ and no closer. Equation~\eqref{eq:pcs} is simply the familiar “favorable/total” ratio, which produces the smooth penalty term in Equation~\eqref{oracle} and seamlessly bridges the two deterministic Cases~1 and~2. Once \(\mathbf{M}_t\) has been constructed, we apply the Hungarian algorithm to \(\mathbf{M}_t\) to obtain \(\mathcal{A}^{\partc}_t\), yielding the optimal \participant{}–spot assignment under this cost structure.

\begin{algorithm}
\caption{Coordinated Oracle (\cordOracle{})}
\label{alg:cord-oracle}
\begin{algorithmic}[1]
 \State \textbf{Input:} $D_t$~—~\participants{} active at step $t$ (incl. unparked \participants{} from $t-1$) \\ 
 $C_t$~—~\competitors{} at time step $t$ (incl. unparked \competitors{} from $t-1$) \\
           $\mathcal{S}_t$~—~available spots at step $t$ \\
           $\{\tau(d,s)\}_{d\in D_t, s\in\mathcal{S}_t}$— all the travel times of \participants{} $d$ to all the spots $s$ 
    \State \textbf{Output:} $\mathcal{A}^{\partc}_t$~—~assigned spots to $D_t$
    \State $\mathcal{A}^{\partc}_t= []$
\State \textbf{Initialize} $\mathbf{M}_t(D_t,\mathcal{S}_t)$
\For{$d \in D_t$}
    \For{$s \in \mathcal{S}_t$}
        \State $\mathbf{M}_t(d,s)= \mathbf{CalculateCost}(\tau(d,s),C_t,R) $ \Comment{\textcolor{blue}{According to Equation~\ref{oracle}.}}
    \EndFor
\EndFor
\State $\mathcal{A}^{\partc}_t = \mathbf{HungarianAssign}$($\mathbf{M}_t$) 
\State \textbf{return} $\mathcal{A}^{\partc}_t$
\end{algorithmic}
\end{algorithm}

\subsubsection{Strategy 4: Coordinated Approximate (\cordApprox{})} Due to the infeasibility of obtaining real-time \competitor{} locations, the \cordOracle{} strategy is impractical in real-world settings.
 The \cordAgn{} strategy, while feasible, suffers from suboptimal allocations as it lacks awareness of \competitors{}. Geographic proximity is a poor proxy for payoff because a “close” space may already be on many \participants{}' radar and disappear before arrival. To account for this hidden competition, we fold in an empirically learned success ratio. We introduce the \cordApprox{} strategy, which uses historical data to probabilistically estimate the behavior of \participants{} and \competitors{}, enhancing decisions without live \competitor{} tracking. This probability modulates travel time, penalizing spots that often snap up quickly and favoring those that are more likely to remain available. The model implicitly captures \participant{} rivalry without simulating each potential \competitor{} explicitly: the lower the historical parking success ratio, the greater the implied crowd pressure. The resulting effective distance steers \participants{} toward spots that are not only near but realistically attainable, reducing wasted searching time and smoothing street parking demand. Let $\mathcal{H} = \{(g_k, \rho_{k,t},t)\}$ be a historical record of occupancy probabilities, where $\rho_{k,t}$ denotes the observed success ratio of finding a spot in Geohash $g_k$ at time $t$. Each cell $z$ in the simulation grid is associated with a Geohash string $g_k$, where $k$ is a unique integer index assigned to that cell for ease of lookup. A forecasting model generates $\hat{p}_{k(s),t}$, the estimated probability of availability in $g_k$ at time $t$. The cost matrix is weighted inversely by success likelihood
\begin{equation}
\tilde{\tau}(d,s) = \dfrac{\tau(d,s)}{\hat{p}_{k(s),t}}
\end{equation}
where $k(s)$ returns the index $k$ of the Geohash $g_k$ containing spot $s$. This effective distance inflates travel time by the inverse of the spot-success prediction. The resulting cost matrix is passed to the Hungarian assignment algorithm, which computes a one-to-one \participant{}-spot match that minimizes the total effective distance. By favoring spaces that are both close and likely to remain free, the scheme reduces futile detours and increases overall parking success compared to distance-only matching. This encodes probabilistic \competitors{} awareness and bridges the gap between the \cordAgn{} (no awareness) and \cordOracle{} (perfect awareness) extremes, achieving near-optimal efficiency while remaining practical. Empirical results in Section~\ref{results} show that \cordApprox{} achieves, on average, 77.54\% of the \participants{} success ratio attained by \cordOracle{} (85.32\%). To estimate spot availability probabilities $\hat{p}_{k(s),t}$, we use Ridge Regression (L2-regularized linear regression)~\cite{hoerl1970ridge}, a robust method for high-dimensional spatiotemporal data with multicollinearity. It mitigates overfitting by penalizing large coefficients, optimizing
\begin{equation}
\hat{\beta} = \arg\min_{\beta} \left\| \mathbf{y} - \mathbf{X} \beta \right\|_2^2 + \lambda \left\| \beta \right\|_2^2
\end{equation}
where $\mathbf{X}$ contains time-of-day, weekday, Geohash (location), and recent occupancy trend features, $\mathbf{y}$ is the historical availability, and $\lambda$ controls regularization strength via cross-validation. This yields stable predictions under sparsity and noise~\cite{hoerl1970ridge, murphy2012machine}. Assignments are logged and merged online, so each round trains on up-to-date observations. This online update loop lets the success ratio estimator track non-stationary demand and gradually refine its forecasts, enabling the assignment policy to converge toward optimal performance. For a detailed description of the procedure, see Algorithm~\ref{alg:agnostic_parking}.

\begin{algorithm}
\caption{Coordinated Approximate (\cordApprox{})}
\label{alg:agnostic_parking}
\begin{algorithmic}[1] 
\State \textbf{Input:} $D_t$~—~\participants{} active at step $t$ (incl. unparked \participants{} from $t-1$) \\ 
           $\mathcal{S}_t$~—~available spots at step $t$ \\
           $\{\tau(d,s)\}_{d\in D_t, s\in\mathcal{S}_t}$~—~all the travel times of \participants{} $d$ to all the spots $s$ \\
           $\mathcal{H} = \{(g_k,\rho_{k,t},t)\}$ — historical record of occupancy probabilities
\State \textbf{Output:} $\mathcal{A}^{\partc}_t$~—~assigned spots to $D_t$
\State $\mathcal{A}^{\partc}_t= []$

\State \textbf{Initialize} $\mathbf{M}_t(D_t,\mathcal{S}_t)$
\For{$d \in D_t$}
    \For{$s \in \mathcal{S}_t$}
        \State $\hat{p}_{k(s),t} \gets \mathbf{Predict}(\mathcal{H}, k(s), t)$
        \State $\tilde{\tau}(d,s)=\frac{\tau(d,s)}{\hat{p}_{k(s),t}}$
        \State $\mathbf{M}_t(d,s) \gets \tilde{\tau}(d,s)$
    \EndFor
\EndFor
\State $\mathcal{A}^{\partc}_t = \mathbf{HungarianAssign}$($\mathbf{M}_t$) 
\State Update $\mathcal{H}$ with new occupancy data
\State \textbf{return} $\mathcal{A}^{\partc}_t$
\end{algorithmic}
\end{algorithm}

\subsection{Data and Simulation Design}
\label{design}
\begin{figure}
  \centering
  \begin{minipage}[t]{0.6\columnwidth}
    \includegraphics[scale=0.079]{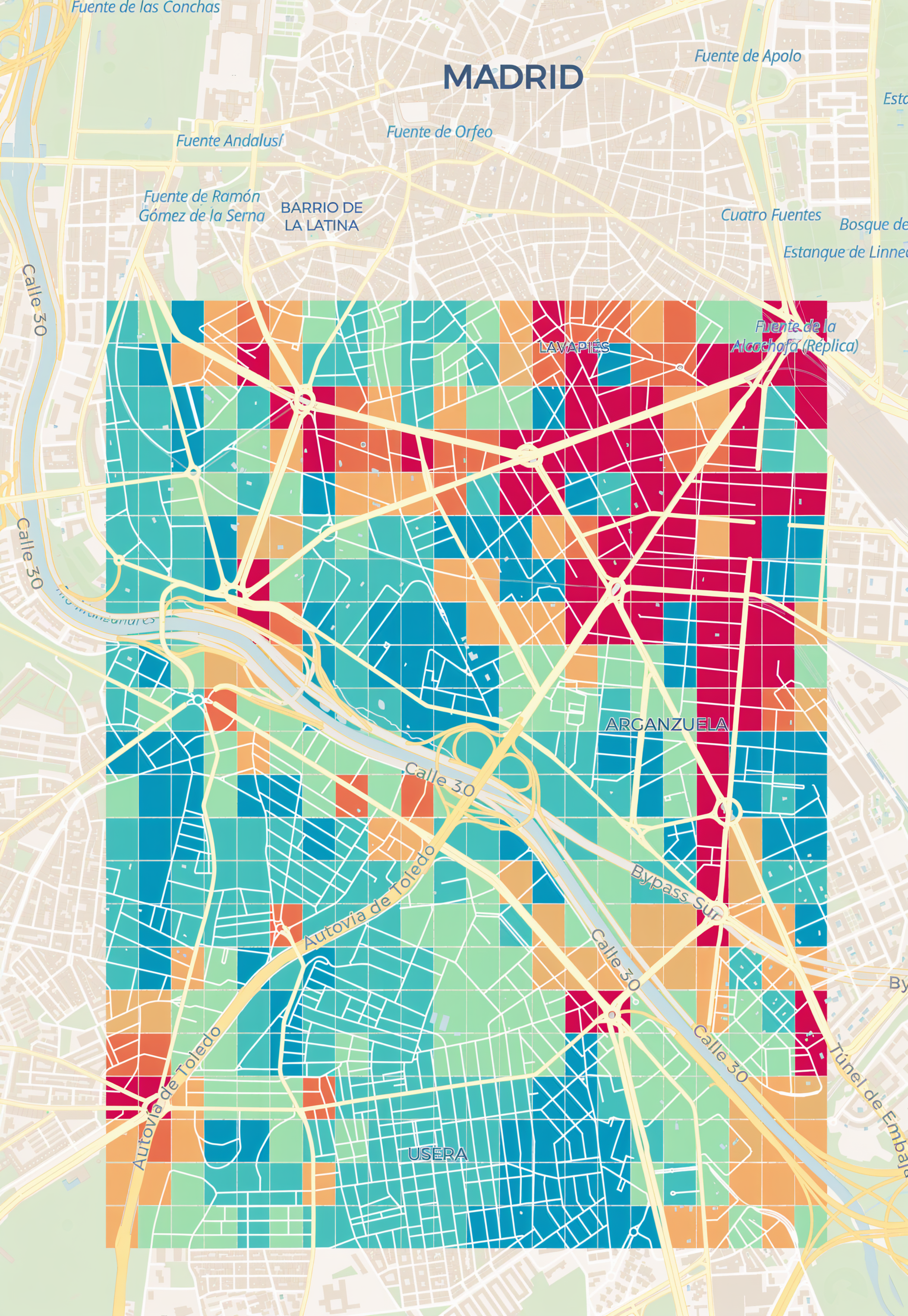}
  \end{minipage}%
  \hfill
    \raisebox{5.8\baselineskip}[0pt][0pt]{
  \begin{minipage}[t]{0.31\columnwidth}
  \centering
  \scriptsize
  \setlength{\tabcolsep}{0.2pt}
  \renewcommand{\arraystretch}{1.1}
  \textbf{Avg.\ Intensity}\\[2pt]
  \begin{tabular}{@{}l@{\,}l@{}}
    \colorbox[HTML]{D50255}{\rule{0pt}{1.5ex}\rule{1em}{0pt}} & 4,025–4,830 \\
    \colorbox[HTML]{EB7053}{\rule{0pt}{1.5ex}\rule{1em}{0pt}} & 3,220–4,025 \\
    \colorbox[HTML]{F5B272}{\rule{0pt}{1.5ex}\rule{1em}{0pt}} & 2,415–3,220 \\
    \colorbox[HTML]{9CE3B1}{\rule{0pt}{1.5ex}\rule{1em}{0pt}} & 1,610–2,415\\
    \colorbox[HTML]{42C1BC}{\rule{0pt}{1.5ex}\rule{1em}{0pt}} & 850–1,610 \\
    \colorbox[HTML]{0198BD}{\rule{0pt}{1.5ex}\rule{1em}{0pt}} & 0–805\\
  \end{tabular}%
  \hspace{3pt}%
\end{minipage}
}
  \caption{Geospatial distribution of traffic intensity across central Madrid, April 18, 2024 (09{:}00–17{:}00).}
  \label{fig:MadridMapIntensity}
\end{figure}

To capture Madrid’s traffic dynamics, we developed a custom simulator from scratch, implementing all algorithms and components ourselves. We first retrieved fine‑grained traffic‑intensity measurements released by the Community of Madrid, Spain~\cite{Dataset}. These counts, recorded at the street‑segment level, capture the temporal and spatial variability of vehicle flow in the city center. Using this dataset, we overlaid a $22 \times 22$ geospatial grid where each cell corresponds to a 7-character Geohash (\(\approx 152.8\,\mathrm{m} \times 116.4\,\mathrm{m}\)). The grid aligns with the cadastral boundaries of central Madrid, encompassing residential neighborhoods and dense commercial corridors (Figure~\ref{fig:MadridMapIntensity}). This granularity allows us to model the heterogeneous distribution of on‑street parking and link each street’s traffic intensity to its enclosing Geohash cell(s). The simulator inherits the “pulse” of the real city: streets that are busy in the empirical record remain busy in the synthetic world, and quiet residential lanes stay quiet. We use a mesoscopic 7-character Geohash grid simulator because the goal is assignment under information limits, not lane dynamics. Coupling to SUMO/MATSim would add overhead without changing the mechanism: the four strategies differ only in how they build the cost matrix passed to the Hungarian algorithm. The grid lets us run 24-h, city-wide, multi-seed sweeps and ablation studies while staying consistent with sensor statistics. We checked correctness on toy cases for Hungarian and by matching simulated diurnal flows to the sensor profile. A microscopic model is only needed for lane-level effects (spillback, signals), which are out of scope. Figures~\ref{fig:MadridMapIntensity}, ~\ref{fig:R1_avg_heatmap},~\ref{fig:main_heatmap_comparison}, and ~\ref{fig:singlecol_heatmap_comparison} are produced with \texttt{kepler.gl}. Basemap data \textcopyright\ OpenStreetMap contributors; tile layer \textcopyright\ CARTO (Voyager).
\\
\noindent $\bullet$ \textbf{Geospatial Data Extraction:} We extract the latitude and longitude coordinates corresponding to each Geohash and map them onto our simulation grid. (Geohashes derived from~\cite{geohashes}).
\\
\noindent $\bullet$ \textbf{Spot Identification:} For each 7-character Geohash, we calculated the maximum on-street parking capacity by summing legally parkable street lengths, accounting for zoning, road class, and municipal restrictions (loading zones, no-stopping, etc.). In simulation, agents park and depart based on real-world dwell times, dynamically updating each Geohash cell's spot availability.
We adopt an extreme-case baseline because producing a complete street space inventory city-wide remains elusive. Existing options, sensor-equipped stalls~\cite{zimmerman2014san}, overhead vision~\cite{zambanini2020detection}, mobile crowd-sensing~\cite{parksense2013}, and crowdsourced maps~\cite{osmParking2022}, are partial, costly, or uneven, so no method yet yields a definitive count. Our one-time manual street length audit, meter-accurate for Madrid’s core, thus provides a reproducible supply ceiling that any empirical estimate will lie below.
\\
\noindent $\bullet$ \textbf{Traffic Intensity Data Extraction:} We extract traffic intensity records for each Geohash cell over a 24-hour period using publicly available data from traffic sensors distributed across Madrid \cite{Dataset}. These sensors record a count each time a vehicle passes, reporting aggregated totals at 15-minute intervals, each linked to a specific street segment. To align this data with our simulation framework, we first map each street to its corresponding Geohash cell(s) based on geographic overlap. These counts are then proportionally assigned to the overlapping Geohash according to the fraction of each street segment contained within each cell. To achieve minute-level granularity, we disaggregate the 15-minute counts into one-minute bins via uniform split, generating a continuous 24-hour intensity profile for each Geohash. Finally, we estimate the share of counts in each cell and time interval that are actively searching for parking, yielding a fine-grained approximation of parking demand intensity at the minute level. Following FHWA guidance, we assume $\approx$10\% of vehicles are searching for parking~\cite{fhwaCruising2023}. Accordingly, we model 1.5\% as \participants{} and 8\% as \competitors{}; the remaining 90.5\% (through-traffic) is outside our simulation scope. We hold adoption shares fixed across strategies to isolate the effect of information and coordination. Across the 24-hour window, the simulator instantiates a total of $\approx$270{,}000 \participants{} (app users) rides and $\approx$1{,}600{,}000 \competitors{} rides (non-users), dispatched hour by hour from the empirical arrival series. The system comprises 12,365 fixed on-street parking spots, distributed across 484 Geohash cells. To reduce stochastic variation, each strategy was run three times.

\section{Experimental Results}
This study models traffic conditions for the week of 15–21 April 2024. We focus on the five working days (Monday 15 – Friday 19), when intensity is relatively stable. Each street segment in the dataset was assigned to a 7-character Geohash \cite{geohashes}, yielding a 22 × 22 lattice (484 Geohashes) that preserves spatial adjacency. Observed traffic counts were projected onto this grid. 18 April 2024 was chosen as the main evaluation date because it is mid-week, avoids holidays, and reflects typical demand. \cordApprox{} was trained on 15–17 April and tested on 18 April. A parallel weekend experiment was conducted on April 20–21 using the \uncAgn{}, \cordAgn{}, and \cordOracle{} strategies. Weekend demand is low-pressure; in that regime \cordAgn{} closely tracks the oracle bound, so we omit \cordApprox{} for brevity. Due to the lower traffic during that period, this experiment is discussed only in Section \ref{zoom}.

We present a 24-hour simulation-based analysis of street parking dynamics within Madrid’s central Geohash grid, where \competitors{} operate within a radius $R=1$; larger $R$ values were tested but are excluded: since at $\approx 152.8{\times}116.4$ m per cell, broader visibility is unrealistic for street parking.  We evaluate how varying levels of information-sharing and coordination impact \participants{}’ ability to successfully secure street parking. First, we compare overall parking success ratios and average search times for four strategies: \uncAgn{}, \cordAgn{}, \cordApprox{}, and \cordOracle{}. Second, we examine how each strategy widens the gap between \participants{} and \competitors{} as availability and demand shifts during the day. Finally, we present a spatial breakdown of performance across land-use zones: commercial, residential, and mixed, on the working day of 18 April 2024, to highlight where \cordApprox{} yields the greatest reductions in search time. We then analyze the same zones under the \cordAgn{} strategy during the weekend (21 April 2024) to examine how different urban areas respond under lower traffic conditions typical of non-working days. These analyses reveal where coordination via historical and real-time data matters most. 

\subsection{City-wide Results}
\label{results}
\subsubsection{Comparative Analysis of Parking Success Ratios Across Strategies and Time Periods}
\begin{figure}[htb]
  \centering
  \begin{subfigure}[b]{0.46\textwidth}
    \includegraphics[page=1,width=\linewidth]{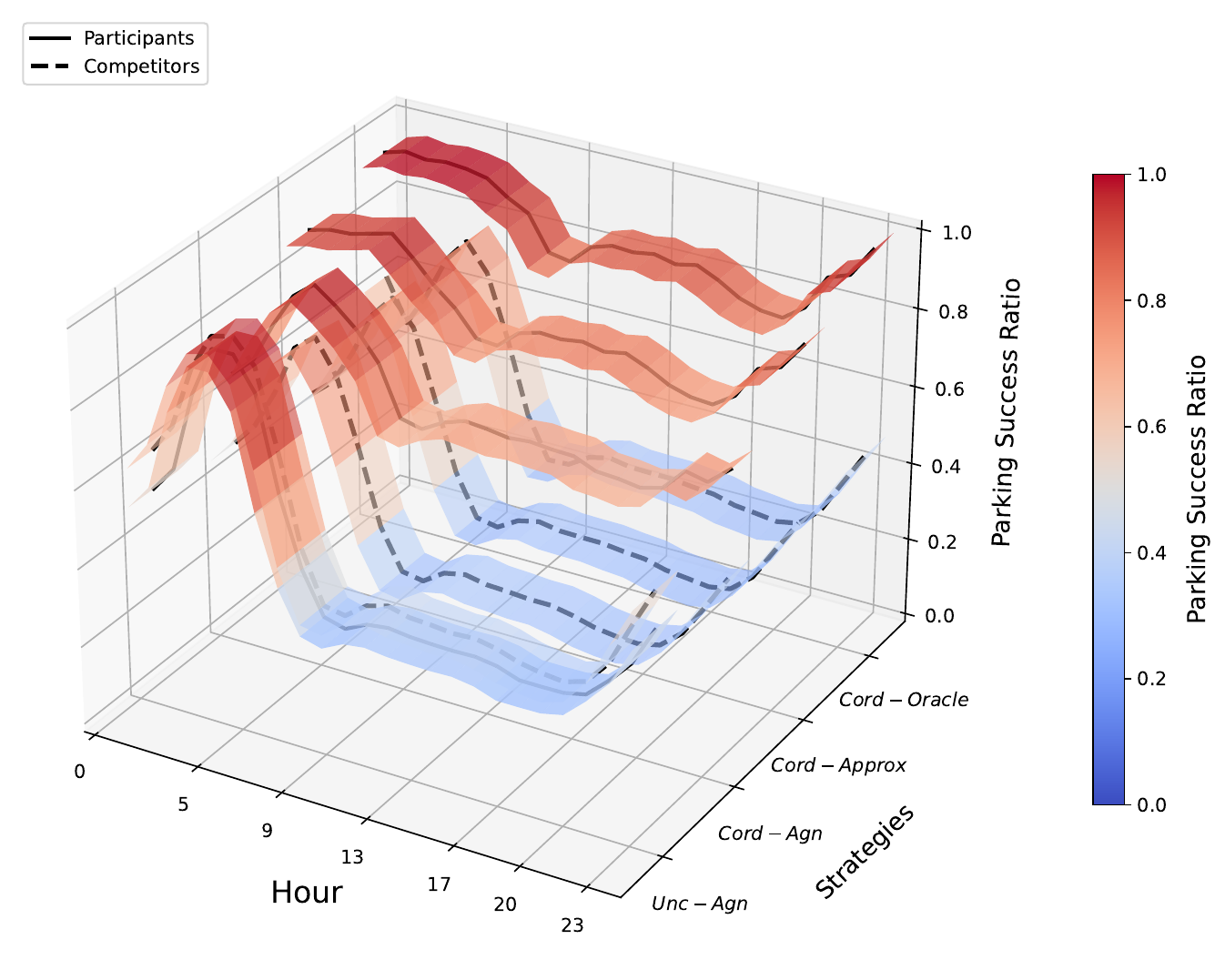}
    \caption{\Participants{} and \competitors{} parking success ratios across strategies over 24 hours. Coordinated approaches significantly improve outcomes during peak hours.}
    \label{subfig:R1_perf}
  \end{subfigure}\hfill
  \begin{subfigure}[b]{0.46\textwidth}
    \includegraphics[page=2,width=\linewidth]{Finalsuccess_merged.pdf}
    \caption{Gap in parking success ratios between \participants{} and \competitors{} by strategies. \cordApprox{} yields the largest gap for \participants{} during peak hours.}
    \label{subfig:R1_diff}
  \end{subfigure}
  \caption{Impact of coordination and information-sharing strategies on parking success ratio.}
  \label{fig:R1_success_ratio}
\end{figure}

Figure \ref{subfig:R1_perf} shows full working day parking success ratios for \participants{} and \competitors{}. During peak hours (09{:}00–17{:}00), \uncAgn{} yields a 34.25\% success ratio for \participants{}, slightly below \competitors{} due to uncoordinated \participants{} converging on the same spot. Coordinated strategies, \cordAgn{}, \cordApprox{}, and \cordOracle{}, raise success ratios to 70.04\%, 77.54\%, and 85.32\%, respectively, achieving gains of 35.79 to 51.07 percentage points (pp) over \uncAgn{} strategy. These improvements stem from more efficient \participants{} distribution and reduced contention. Notably, \cordApprox{} closely approaches \cordOracle{}, showing that historical data can approximate \competitors{} effects without real-time tracking. The 35.79 pp gap between \cordAgn{} and \uncAgn{} illustrates that coordination alone offers greater benefits than spot-level information. During off-peak hours, differences are narrow, as low competition allows uninformed \competitors{} to park successfully. 
\begin{table}[h!]
  \caption{Average success ratio and search time (09{:}00–17{:}00).}
  \centering
  \resizebox{\columnwidth}{!}{%
  \begin{tabular}{lcccc}  
    \toprule
    & \multicolumn{2}{c}{\textbf{Parking Success Ratio (\%)}} 
    & \multicolumn{2}{c}{\textbf{Average Search Time (min)}} \\
    \cmidrule(lr){2-3}\cmidrule(lr){4-5}
    \textbf{Strategies} & \textbf{\Participants{}} & \textbf{\Competitors{}} 
                      & \textbf{\Participants{}} & \textbf{\Competitors{}} \\
    \midrule
    1 \uncAgn{}      & 34.25 & 38.63 & 19.24 & 17.83 \\
    2 \cordAgn{}     & 70.04 & 30.71 &  9.86 & 19.79 \\
    3 \cordApprox{}   & 77.54 & 28.71 &  \textbf{6.69} & 19.98 \\
    4 \cordOracle{}  & \textbf{85.32} & 27.76 &  8.83 & 19.89 \\
    \bottomrule
  \end{tabular}}
  \label{table}
\end{table}
Figure \ref{subfig:R1_diff} illustrates how different parking strategies affect the success gap between \participants{} and \competitors{}. Counter-intuitively, using only information about parking alone can actually harm the \participants{} (-4.38 pp). This occurs because \uncAgn{} may direct multiple \participants{} to the same spot, leaving most without parking; when frequent, random searching can outperform informed but uncoordinated choices. Coordinated strategies yield substantial improvements in \participants{} parking success ratios: \cordAgn{} achieves 70.04\% (+39.33 pp), \cordApprox{} reaches 77.54\% (+48.83 pp), and \cordOracle{} attains 85.32\% (+57.56 pp), all relative to \competitors{}. These results confirm that while information helps, coordination is essential for parking success ratio (see Table \ref{table}). However, \participants{} who couldn’t find parking within $t_{max} = 30$ minutes were effectively “forced out”, they either had to park farther away and arrange alternate transport back to their destination, or pay for costly private parking.

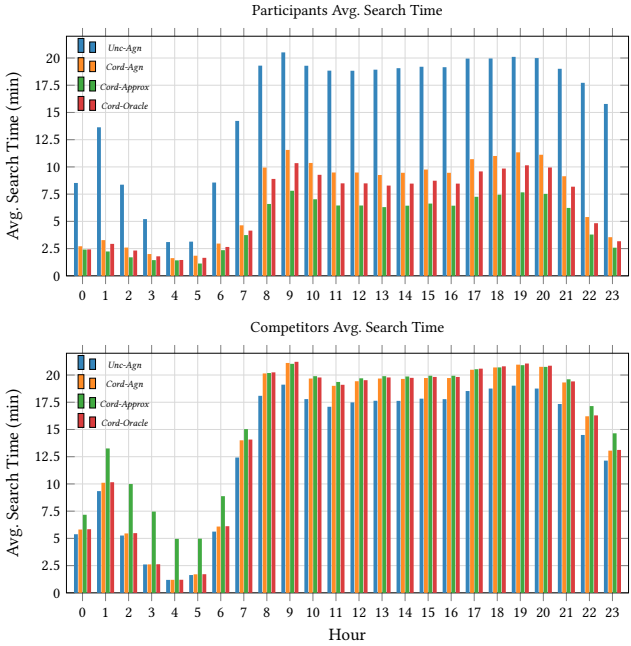
\begin{figure}[h!]
  \centering
  \resizebox{1\linewidth}{!}{
    \begin{tikzpicture}
      \begin{groupplot}[
        group style={
          group size=1 by 2,
          vertical sep=40pt
        },
        width=1.39\linewidth, 
        height=0.7\linewidth, 
        ybar,
        ymin=0, ymax=22,
        ytick distance=2.5,
        ymajorgrids=true,
        xmajorgrids=true,
        grid style={solid, thin, draw=gray!30},
        tick label style={font=\small},
        xlabel style={font=\normalsize},
        title style={font=\small}
      ]
        \nextgroupplot[
          title={\Participants{} Avg.\ Search Time},
          ylabel={Avg.\ Search Time (min)},
          bar width=0.07cm,
          xmin   = -3,
          xmax   = 23*4.3 + 3,
          enlarge x limits=false,
          xtick={
            0*4.3, 1*4.3, 2*4.3, 3*4.3, 4*4.3, 5*4.3,
            6*4.3, 7*4.3, 8*4.3, 9*4.3, 10*4.3, 11*4.3,
            12*4.3,13*4.3,14*4.3,15*4.3,16*4.3,17*4.3,
            18*4.3,19*4.3,20*4.3,21*4.3,22*4.3,23*4.3
          },
          xticklabels={0,1,2,3,4,5,6,7,8,9,10,11,12,13,14,15,16,17,18,19,20,21,22,23},
          legend style={
            at={(0.009,1.01)}, anchor=north west,
            legend columns=1, draw=none, fill=none,
            font=\footnotesize
          }
        ]
          \addplot[draw=none,fill=colorUncAgn,   bar shift=-0.12cm]
            table[x expr=\coordindex*4.3, y=greedy_drv_avg]   {\datatable};
          \addplot[draw=none,fill=colorCordAgn,  bar shift=-0.04cm]
            table[x expr=\coordindex*4.3, y=agnostic_drv_avg]{\datatable};
          \addplot[draw=none,fill=colorCordAprox,bar shift=0.04cm]
            table[x expr=\coordindex*4.3, y=cutoff_drv_avg]  {\datatable};
          \addplot[draw=none,fill=colorCordOracle,bar shift=0.12cm]
            table[x expr=\coordindex*4.3, y=oracle_drv_avg]  {\datatable};
          \legend{%
            {\tiny\uncAgn{}},
            {\tiny\cordAgn{}},
            {\tiny\cordApprox{}},
            {\tiny\cordOracle{}}%
          }

        \nextgroupplot[
          title={\Competitors{} Avg.\ Search Time},
          ylabel={Avg.\ Search Time (min)},
          bar width=0.07cm,
          xlabel={Hour},
          xmin   = -3,
          xmax   = 23*4.3 + 3,
          enlarge x limits=false,
          xtick={
            0*4.3, 1*4.3, 2*4.3, 3*4.3, 4*4.3, 5*4.3,
            6*4.3, 7*4.3, 8*4.3, 9*4.3, 10*4.3, 11*4.3,
            12*4.3,13*4.3,14*4.3,15*4.3,16*4.3,17*4.3,
            18*4.3,19*4.3,20*4.3,21*4.3,22*4.3,23*4.3
          },
          xticklabels={0,1,2,3,4,5,6,7,8,9,10,11,12,13,14,15,16,17,18,19,20,21,22,23},
          legend style={
            at={(0.009,1.01)}, anchor=north west,
            legend columns=1, draw=none, fill=none,
            font=\footnotesize
          }
        ]
          \addplot[draw=none,fill=colorUncAgn,   bar shift=-0.12cm]
            table[x expr=\coordindex*4.3, y=greedy_comp_avg]   {\datatable};
          \addplot[draw=none,fill=colorCordAgn,  bar shift=-0.04cm]
            table[x expr=\coordindex*4.3, y=agnostic_comp_avg]{\datatable};
          \addplot[draw=none,fill=colorCordAprox,bar shift=0.04cm]
            table[x expr=\coordindex*4.3, y=cutoff_comp_avg]  {\datatable};
          \addplot[draw=none,fill=colorCordOracle,bar shift=0.12cm]
            table[x expr=\coordindex*4.3, y=oracle_comp_avg]  {\datatable};
          \legend{%
            {\tiny\uncAgn{}},
            {\tiny\cordAgn{}},
            {\tiny\cordApprox{}},
            {\tiny\cordOracle{}}%
          }
      \end{groupplot}
    \end{tikzpicture}
  }
  \caption{Avg. parking search time (min) for \participants{} and \competitors{} across strategies, with coordinated \participants{} seeing notable reductions during peak hours 09{:}00–17{:}00.}
  \label{fig:R1_search_times_driver}
\end{figure}
\subsubsection{Comparative Analysis of Average Search Time Across Strategies and Time Periods}
Figure~\ref{fig:R1_search_times_driver} presents the average \participant{} search time across the 24-hour duration of the working day. Table \ref{table} complements this view by presenting peak‑hour (09{:}00–17{:}00) search time for \participants{} and \competitors{}. We benchmark each strategy against the \uncAgn{} strategy and evaluate two objectives: (i) reducing \participants{} search time and (ii) widening the gap over \competitors{}. As expected, \uncAgn{} performs worst: \participants{} require on average 19.24 minutes to locate a spot, marginally more than \competitors{} at 17.83 minutes, confirming the poor parking success ratio reported in Figure \ref{subfig:R1_perf}. Adding coordination substantially improves performance. Under \cordAgn{}, \participants{} search time drops to 9.86 minutes, a 48.7\% reduction relative to \uncAgn{}, while \competitors{} time increases to 19.79 minutes. \cordApprox{} further reduces \participant{} search time to 6.69 minutes (65.2\%) and raises \competitors{} time to 19.98 minutes. Finally, \cordOracle{} yields 8.83 minutes for \participants{} (54.1\%) and 19.89 minutes for \competitors{}. The fact that \cordApprox{} achieves shorter search time than \cordOracle{} is not unexpected. The metric is computed only for successful parking events, and the two strategies differ in success ratio; therefore the values are not directly comparable. Overall, \cordApprox{} delivers the best trade‑off, cutting \participants{} search time by 66.5\% relative to \competitors{} and by 65.2\% with respect to the baseline of the \participants{}, \uncAgn{} strategy.
\subsubsection{Comparative Analysis of Agents’ Parking Success Ratios Across Strategies Under Varying Parking Spot Availability.} 
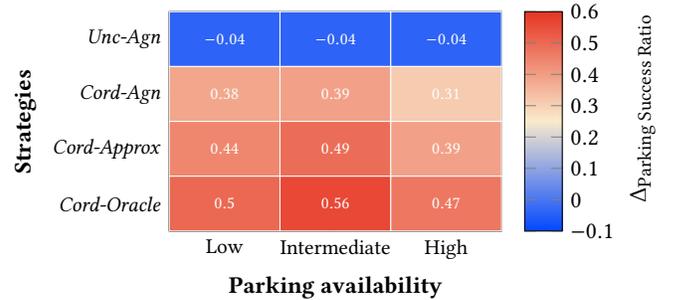
\begin{figure}[htb]
\centering
\begin{tikzpicture}
  \begin{axis}[
    width=6cm,
    height=4.5cm,
    colormap={purplewhitered}{
      rgb(0cm)=(0.01,0.28,1);
      rgb(1cm)=(0.98,0.92,0.8);
      rgb(2cm)=(0.9,0.2,0.13)
    },
    colorbar,
    colorbar style={
      ylabel={$\Delta_{\text{Parking Success Ratio}}$},
      ylabel style={
        rotate=360,
        anchor=north,    
        xshift=5pt, 
      },
      ytick={-0.1,0,0.1,0.2,0.3,0.4,0.5,0.6},
      ymin=-0.1,
      ymax=0.6,
    },
    point meta min=-0.1,
    point meta max=0.6,
    enlargelimits=false,
    xtick={0,1,2},
    ytick={0,1,2,3},
    xticklabels={{\small Low},{\small Intermediate},{\small High}},
    yticklabels={
      {\small\uncAgn{}},
      {\small\cordAgn{}},
      {\small\cordApprox{}},
      {\small\cordOracle{}}
    },
    y dir=reverse,
    xlabel={\textbf{Parking availability}},
    ylabel={\textbf{Strategies}},
    nodes near coords,
    every node near coord/.append style={
      font=\scriptsize,
      align=center,
      anchor=center,
      text=white,
      /pgf/number format/fixed,
      /pgf/number format/precision=2
    },
  ]
    \addplot[
      matrix plot*,
      mesh/cols=3,
      draw=white,
      point meta=explicit
    ] table[meta=c] {
      x y c
      0 0 -0.04
      1 0 -0.04
      2 0 -0.04
      0 1 0.38
      1 1 0.39
      2 1 0.31
      0 2 0.44
      1 2 0.49
      2 2 0.39
      0 3 0.5
      1 3 0.56
      2 3 0.47
    };
  \end{axis}
\end{tikzpicture}
\caption{Difference in avg. parking success ratios between \participants{} and \competitors{} across varying parking availability; low (0–5\%), intermediate (20–25\%), and high (40–45\%).}
\label{fig:gap_heatmap}
\end{figure}

Figure~\ref{fig:gap_heatmap} visualizes the average parking success ratio gap in the working day between \participants{} and \competitors{} under different information, coordination strategies, and traffic conditions (i.e., street parking availability). Under high parking availability (40--45\%), both agents perform well, with limited contention for spots, resulting in moderate but consistent success gaps ranging from +0.31 to +0.47. These gaps reflect the residual benefit of coordination, even when parking is relatively abundant. However, under low parking availability (0--5\%), overall success ratios decline sharply, ranging from +0.38 to +0.5, due to fierce competition and spot scarcity. Interestingly, while the success gaps remain slightly wider, the benefits of coordination are diminished, as structural limitations, namely, the severe shortage of available spots, leave little room for any strategy to improve outcomes, regardless of planning. The most informative regime emerges at intermediate availability (20--25\%), where supply and demand are balanced enough to enable effective competition. Here, coordination strategies demonstrate their full potential. All coordinated approaches maintain high parking success ratios (above 70\%), and the \participants{}–\competitors{} gap reaches its maximum, peaking at +0.56 under the \cordOracle{} strategy. This regime highlights the ideal operating point where predictive coordination most effectively differentiates \participants{} outcomes, leveraging limited but actionable information to outperform \competitors{}.
\begin{figure*}
  \centering
    \captionsetup[subfigure]{position=bottom,
                        singlelinecheck=true,
                        justification=centering}
  \begin{minipage}[t]{0.31\textwidth}
    \centering
    \includegraphics[scale=0.0557]{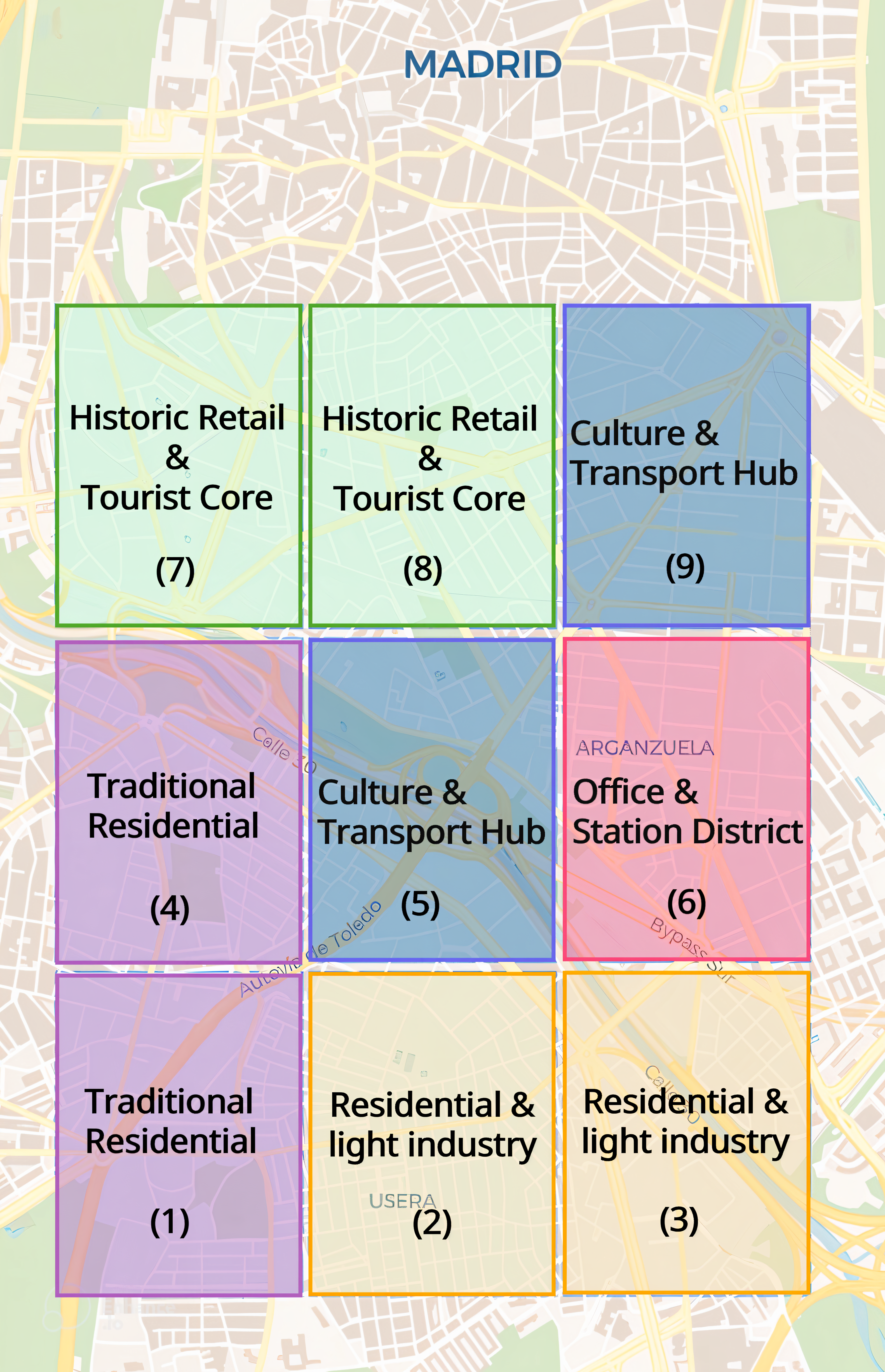}
    \captionof{figure}{A schematic 3×3 grid overlay of the central Madrid study area, showing nine discrete spatial cells classified into five predominant land-use types.}
    \label{fig:R1_avg_heatmap}
  \end{minipage}
  \hfill
  \begin{minipage}[t]{0.65\textwidth}
    \centering
    \begin{subfigure}[t]{0.42\linewidth}
      \centering
      \includegraphics[width=\linewidth]{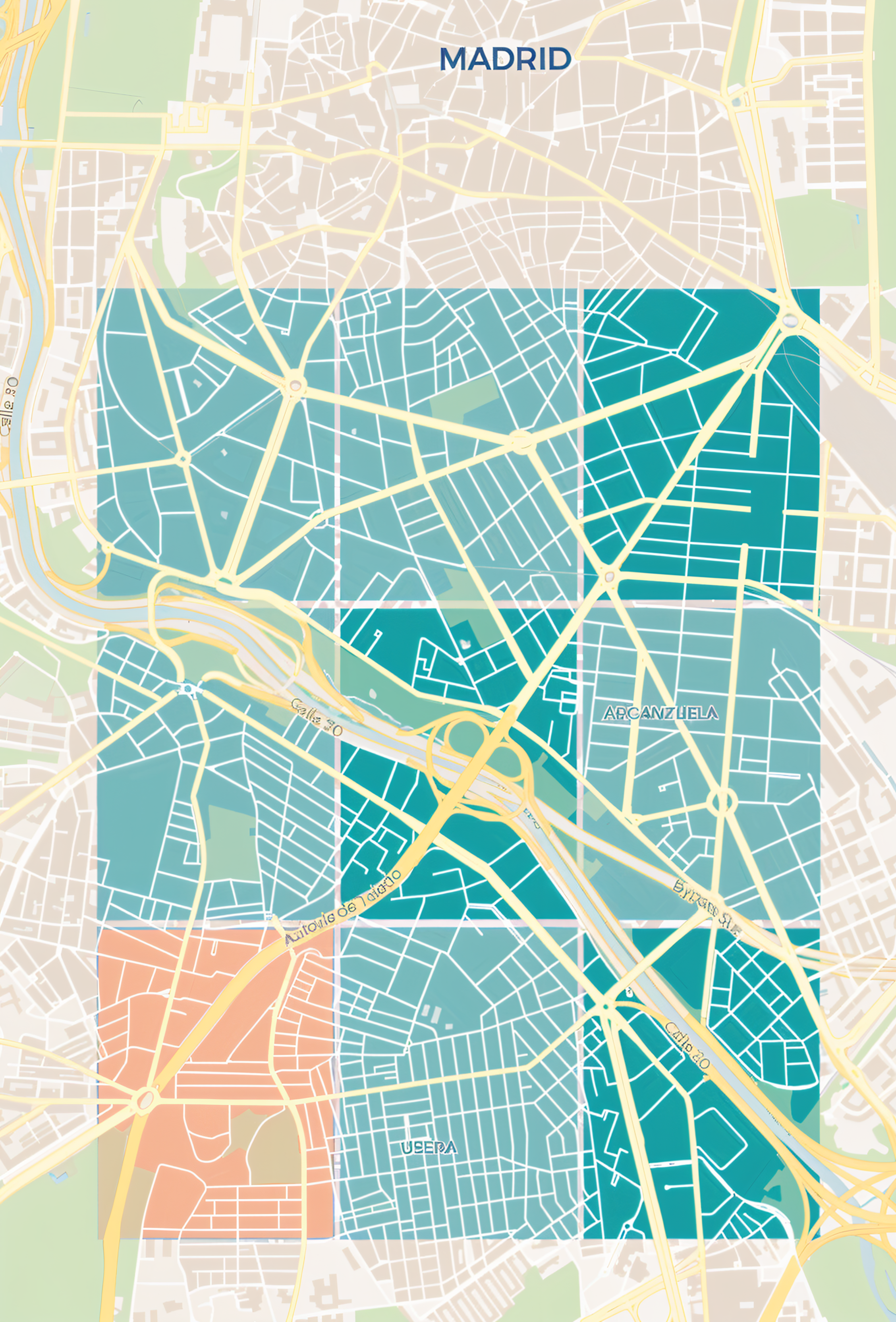}%
      \caption{Avg.\ search time for \participants{}.}
      \label{fig:R2_avg_heatmap}%
    \end{subfigure}
    \hfill
    \begin{subfigure}[t]{0.42\linewidth}
      \centering
      \includegraphics[width=\linewidth,height=1.49\textwidth]{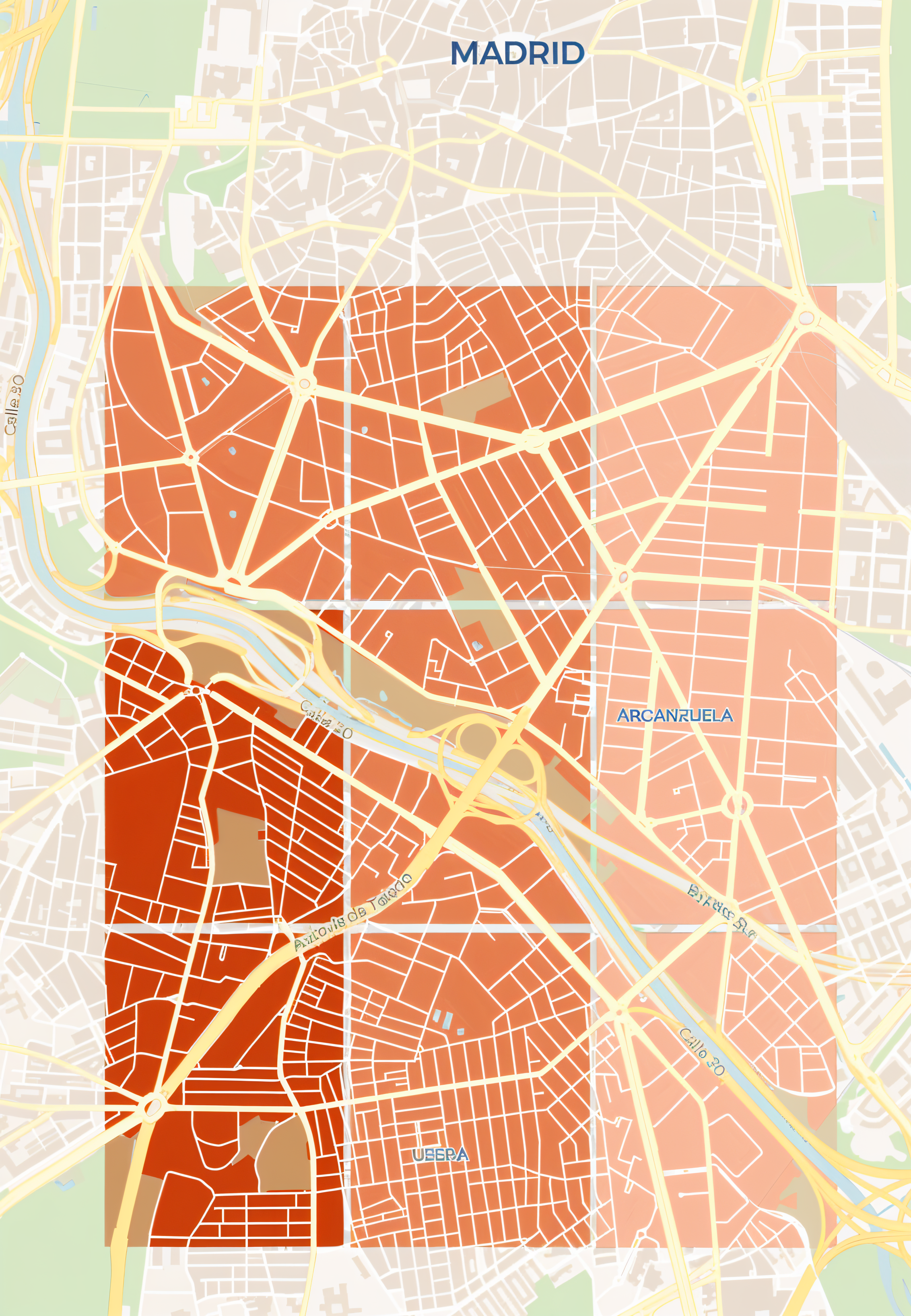}%
      \caption{Avg.\ search time for \competitors{}.}%
      \label{fig:R3_avg_heatmap}%
    \end{subfigure}
    \vspace{3ex} 
    \raisebox{5\baselineskip}[0pt][0pt]{
    \begin{minipage}{0.5\linewidth}
      \centering
      \scriptsize
      \setlength{\tabcolsep}{2pt}
      \renewcommand{\arraystretch}{1.1}
      Avg.\ Search\\Time (min)\\[3pt]
      \begin{tabular}{@{}l@{\,}l@{}}
      \colorbox{redDark!80}{\rule{0pt}{1.5ex}\rule{1em}{0pt}} & 25–28 \\
      \colorbox{peachDark!80}{\rule{0pt}{1.5ex}\rule{1em}{0pt}} & 20–25 \\ 
      \colorbox{peachLight!80}{\rule{0pt}{1.5ex}\rule{1em}{0pt}} & 15–20 \\
      \colorbox{tealLight!80}{\rule{0pt}{1.5ex}\rule{1em}{0pt}} & 10–15 \\
      \colorbox{tealMedium!80}{\rule{0pt}{1.5ex}\rule{1em}{0pt}} & 5–10 \\
      \colorbox{tealDark!80}{\rule{0pt}{1.5ex}\rule{1em}{0pt}} & 0-5 \\ 
      \end{tabular}
    \end{minipage}
    }
    \vspace{1ex}
    \vspace{-2em}
    \captionof{figure}{Working-day peak-hour (09{:}00–17{:}00) average search time under \cordApprox{} in central Madrid: (a) \participants{}, (b) \competitors{}. Each 3×3 grid cell is color-coded by mean search time (min), with a shared legend.}
    \label{fig:main_heatmap_comparison}
  \end{minipage}
\end{figure*}
\subsection{Zooming in on Particular Zones of Madrid}
\label{zoom}
We discretized the central Madrid study area into nine distinct zones, each characterized by one of five predominant land-use archetypes: \textbf{Historic Retail \& Tourist Core}, \textbf{Cultural \& Transport Hub}, \textbf{Traditional Residential}, \textbf{Office \& Station District}, and \textbf{Residential \& Light Industry}. As shown in Figure~\ref{fig:R1_avg_heatmap}, each zone is numbered to facilitate clearer referencing and interpretation throughout our analysis. This spatial categorization allows us to capture the heterogeneity of urban environments and explore how different patterns of land use shape parking behavior and search dynamics. By zooming into this grid-level view, we uncover performance variations that are not evident from city-wide aggregates.

\subsubsection{Working Day Peak Hour (18 April 2024, 09{:}00–17{:}00)}
In Figure~\ref{fig:main_heatmap_comparison}, during peak working-day hours, \participants{} using the \cordApprox{} strategy significantly outperform \competitors{} across all zones, demonstrating the strategic value of coordination and predictive modeling. Table~\ref{tab:search_time_with_averages} reveals how zone-specific traffic form and usage patterns mediate these gains:
\\
\noindent $\bullet$ \textbf{Traditional Residential.}
\Participants{} in Cell (4) achieve a substantial 73\% improvement (7.00 min vs. 26.14 min for \competitors{}), primarily due to moderate density, regular street layouts, and consistent demand patterns that enhance the reliability of predictive coordination. In contrast, Cell (1), featuring narrower roads and higher congestion, yields a lower but still meaningful 40\% improvement (16.69 min vs. 28.01 min), reflecting the limitations of prediction under more constrained spatial conditions.
\\
\noindent $\bullet$ \textbf{Residential \& Light Industry.}
These zones offer predictable demand cycles, such as morning arrivals and afternoon departures tied to shift work and local commerce. Cell (2) shows a 68\% decrease (6.39 min vs. 19.90 min), while Cell (3), with its lower baseline congestion, demonstrates an even stronger 78\% improvement (2.58 min vs. 11.93 min), underscoring how excess capacity amplifies the benefits of strategic guidance.
\\
\noindent $\bullet$ \textbf{Cultural \& Transport Hubs.}
These zones exhibit high turnover due to continuous taxi, tourist, and passenger activity, making real-time spot prediction highly effective. Cell (5) records a 77\% reduction (4.57 min vs. 19.70 min), while Cell (9) achieves a similar 75\% improvement (3.45 min vs. 13.97 min). High spot turnover enhances the effectiveness of \cordApprox{}, particularly when coordinated with known traffic patterns.
\\
\noindent $\bullet$ \textbf{Office \& Station District.}
This zone features strong temporal rhythms due to synchronized train schedules and commuter flows. Cell (6) shows a 49\% improvement (6.66 min vs. 13.17 min), as the predictability of demand surges allows preemptive assignment strategies to operate effectively within structured time windows.
\\
\noindent $\bullet$ \textbf{Historic Retail \& Tourist Core.}
Rapid spot turnover driven by short-term visitors, deliveries, and shopping trips allows predictive strategies to substantially reduce search burden. Cell (7) achieves a 69\% improvement (7.69 min vs. 24.66 min), and Cell (8), despite a more complex and irregular street layout, still yields a notable 64\% reduction (7.07 min vs. 19.82 min), highlighting that turnover frequency can offset geometric inefficiencies in the road network.
\begin{table*}[htb]
  \caption{Average search times (minutes) for \participants{} and \competitors{} across defined zones in central Madrid during peak hours (09{:}00–17{:}00), comparing Working day (\cordApprox{}) and Weekend (\cordAgn{}) strategies.}
  \label{tab:search_time_with_averages}
  \centering
  \begin{tabular}{l  c  c  c  c}
    \toprule
    \multirow{2}{*}{\textbf{Zone}} 
      & \multicolumn{2}{c}{\textbf{Working Day}} 
      & \multicolumn{2}{c}{\textbf{Weekend}} \\
    \cmidrule(lr){2-3} \cmidrule(lr){4-5}
      & \textbf{\Participants{}} & \textbf{  \Competitors{}  } 
      & \textbf{\Participants{}} & \textbf{  \Competitors{}  } \\
    \midrule
    \textbf{Traditional Residential (avg.)}      & 11.84 & 27.08 &  3.81 & 11.60 \\
    Traditional Residential (1)                 & 16.69 & 28.01 &  3.85 & 11.84 \\
    Traditional Residential (4)                 &  7.00 & 26.14 &  3.77 & 11.35 \\
    \addlinespace
    \textbf{Residential \& Light Industry (avg.)}  &  4.48 & 15.91 &  3.67 &  5.41 \\
    Residential \& Light Industry (2)             &  6.39 & 19.90 &  4.17 &  7.31 \\
    Residential \& Light Industry (3)             &  2.58 & 11.93 &  3.17 &  3.52 \\
    \addlinespace
    \textbf{Culture \& Transport Hub (avg.)}       &  4.01 & 16.83 &  4.53 &  6.15 \\
    Culture \& Transport Hub (5)                  &  4.57 & 19.70 &  5.03 &  7.37 \\
    Culture \& Transport Hub (9)                  &  3.45 & 13.97 &  4.03 &  4.93 \\
    \addlinespace
    \textbf{Office \& Station District (6)}       &  6.66 & 13.17 &  7.43 &  4.39 \\
    \addlinespace
    \textbf{Historic Retail \& Tourist Core (avg.)} &  7.38 & 22.24 &  6.73 & 9.64 \\
    Historic Retail \& Tourist Core (7)            &  7.69 & 24.66 &  6.01 & 11.69 \\
    Historic Retail \& Tourist Core (8)            &  7.07 & 19.82 &  7.44 &  7.59 \\
    \bottomrule
  \end{tabular}
\end{table*}
\subsubsection{Weekend Peak Hour (21 April 2024, 09{:}00–17{:}00)}
\begin{figure}[htb]
  \centering
  \captionsetup[subfigure]{position=bottom,
                            singlelinecheck=true,
                            justification=centering}
  \begin{minipage}[t]{0.41\columnwidth}
    \centering
    \includegraphics[width=\linewidth]{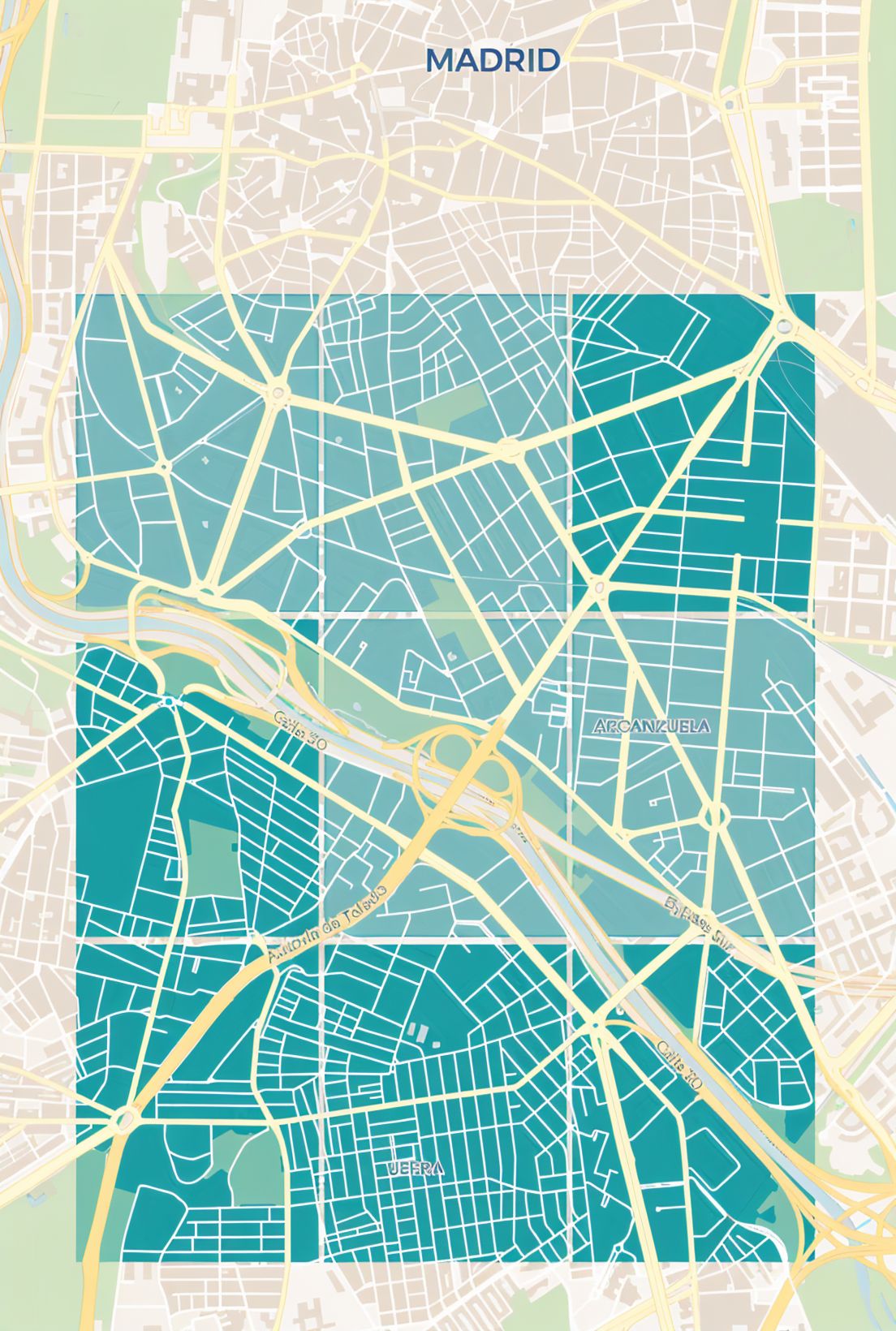}%
    \subcaption{Avg.\ search time for \participants{}.}
    \label{fig:week-driver}
  \end{minipage}%
  \hspace{0.01\columnwidth}
  \begin{minipage}[t]{0.1\columnwidth}
    \centering
    \tiny
    \setlength{\tabcolsep}{0.2pt}
    \renewcommand{\arraystretch}{1.0}
    \raisebox{5.5\baselineskip}[0pt][0pt]{
      \begin{minipage}{\columnwidth}
        \centering
        Avg.\ Search\\Time (min)\\[1pt]
        \begin{tabular}{@{}l@{\,}l@{}}
        \colorbox{redDark!80}{\rule{0pt}{1.5ex}\rule{1em}{0pt}} & 25–28 \\
        \colorbox{peachDark!80}{\rule{0pt}{1.5ex}\rule{1em}{0pt}} & 20–25 \\
        \colorbox{peachLight!80}{\rule{0pt}{1.5ex}\rule{1em}{0pt}} & 15–20 \\
        \colorbox{tealLight!80}{\rule{0pt}{1.5ex}\rule{1em}{0pt}} & 10–15 \\ 
        \colorbox{tealMedium!80}{\rule{0pt}{1.5ex}\rule{1em}{0pt}} & 5–10 \\
        \colorbox{tealDark!80}{\rule{0pt}{1.5ex}\rule{1em}{0pt}} & 0-5 \\ 

        \end{tabular}
      \end{minipage}
      }
  \end{minipage}
  \hspace{0.01\columnwidth}
  \begin{minipage}[t]{0.41\columnwidth}
    \centering
    \includegraphics[width=\linewidth,height=1.49\linewidth]{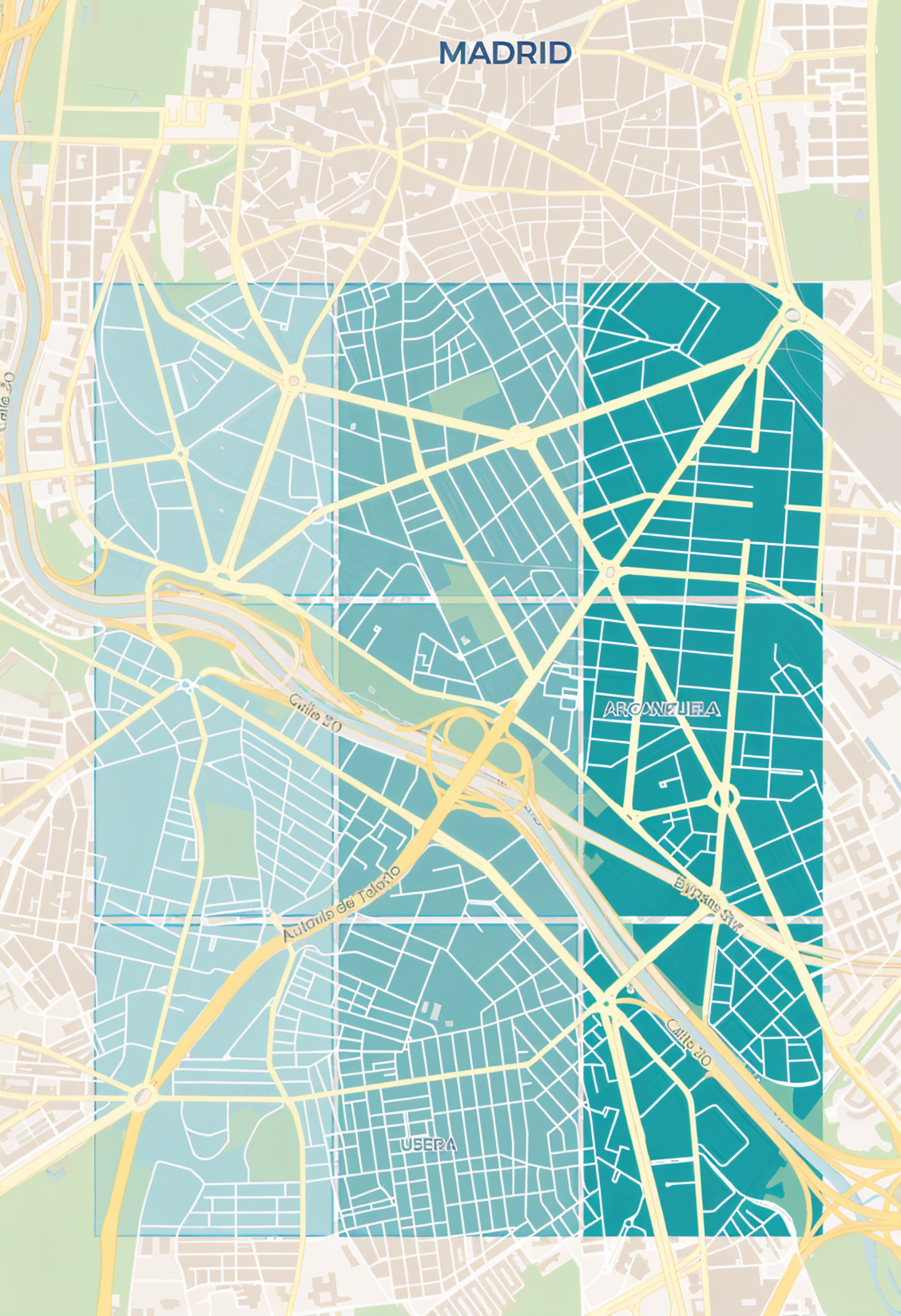}
    \subcaption{Avg.\ search time for \competitors{}.}
    \label{fig:week-comp}
  \end{minipage}
  \vspace{1ex}
  \caption{
    Weekend peak-hour avg. search time under \cordAgn{}: (a) \participants{}, (b) \competitors{}. Each 3×3 grid cell is color‐coded by mean search time (min), with a shared legend.}
  \label{fig:singlecol_heatmap_comparison}
\end{figure}

Weekend conditions, with lower traffic and competition, diminish the benefits of the advanced \cordApprox{} strategy (Figure~\ref{fig:singlecol_heatmap_comparison}). We therefore focus on the \cordAgn{} strategy, which closely mirrors \cordOracle{} performance under these simpler dynamics.
\\
\noindent $\bullet$ \textbf{Traditional Residential.} \Participants{} in these zones experience a consistent 67\% reduction in search time (3.8 min vs. 11.6 min), as weekend travel patterns ease pressure on street parking. With many residents inactive or away, natural availability improves, making even simple coordination highly effective.
\\
\noindent $\bullet$ \textbf{Residential \& Light Industry.} Weekend dynamics yield mixed results. Cell (2) shows a 43\% reduction (4.17 min vs. 7.31 min), likely due to residual working day patterns, even though commercial deliveries are largely inactive. Cell (3), with minimal congestion, sees only marginal gains (3.17 min vs. 3.52 min), showing limited value of coordination in low-pressure settings.
\\
\noindent $\bullet$ \textbf{Culture \& Transport Hubs.} These zones benefit from ongoing but lower-intensity weekend movement, such as leisure trips, museum visits, or local transit use, yielding modest yet consistent gains. Cell (5) improves by 32\% (5.03 min vs. 7.37 min), and Cell (9) by 18\% (4.03 min vs. 4.93 min), suggesting that even light but steady turnover sustains coordination value.
\\
\noindent $\bullet$ \textbf{Office \& Station District.} On the weekend, this area lacks the structured commuter flows seen on the working day. Instead, irregular and unpredictable visitor patterns dominate, leading to weaker \participants{}' outcomes (7.43 min vs. 4.39 min) as \cordAgn{} becomes less effective without consistent demand rhythms.
\\
\noindent $\bullet$ \textbf{Historic Retail \& Tourist Core.} These zones show contrasting behaviors. Cell (7) benefits from continued visitor activity; shopping, dining, and sightseeing, leading to a 49\% improvement (6.01 min vs. 11.69 min). However, Cell (8), where café and restaurant patrons occupy spaces for extended periods, shows almost no gain (7.44 min vs. 7.59 min), highlighting how static parking behavior limits strategy effectiveness.

Comparative analysis of working day and weekend results highlights that \cordApprox{} is particularly effective under working day conditions, where higher traffic demand and competitive pressure necessitate more advanced strategies. On weekends, however, the reduced demand and naturally improved street parking spot availability render such complexity unnecessary. In these cases, simpler coordination approaches suffice, maintaining strong performance without the overhead of approximation.

\section{Conclusion}
We studied how information and coordination shape on-street parking outcomes at city scale, comparing four strategies from information-only (\uncAgn{}) to an oracle full knowledge upper bound (\cordOracle{}). Information alone is not enough: when many participants chase the same spots, \uncAgn{} can harm them, whereas coordination (\cordAgn{}) yields large gains but still trails the oracle. To bridge this gap, we introduced \cordApprox{}, which weights distance by learned success probabilities from historical occupancy and then solves a Hungarian assignment, capturing much of the oracle’s benefit without live tracking of \competitors{} or new sensing. Under weekday operating conditions (with \competitors{} observability fixed at $R{=}1$), \cordApprox{} reduces \participants{}’ average search time to \emph{6.69} minutes vs.\ \emph{19.98} for \competitors{}, with gains peaking at intermediate availability ($\approx$20–25\%) and tapering at extremes. On low-pressure weekends, \cordAgn{} nearly matches the oracle, so approximation adds little. We report both users (\participants{}) and non-users (\competitors{}) outcomes and flag equity/access considerations. Limitations include fixed adoption, $R{=}1$, and search-time computed for successful attempts only. Future work will study adoption (scaling) sensitivity, congestion/VKT and emissions, robustness to imperfect sensing/partial observability (incl.\ $R$ sensitivity), and privacy-preserving or decentralized variants. Overall, \cordApprox{} offers a practical balance: near-oracle benefits with deployable data and no live \competitors{} tracking.

\begin{acks}
This work was supported by the MLEDGE project (TSI-063100-2022-0004), funded by the Ministry of Economic Affairs and Digital Transformation and the European Union NextGenerationEU/PRTR.
\end{acks}

\end{document}